\documentclass[10pt,twocolumn,letterpaper]{article}

\usepackage{wacv}
\usepackage{times}
\usepackage{epsfig}
\usepackage{graphicx}
\usepackage{amsmath}
\usepackage{amssymb}
\usepackage{subcaption}
\usepackage{tabularx}
\usepackage{paralist}
\usepackage[dvipsnames]{xcolor}
\usepackage[symbol]{footmisc}

\def\wacvPaperID{347} 

\wacvfinalcopy 

\ifwacvfinal
\fi

\usepackage[pagebackref=true,breaklinks=true,colorlinks,bookmarks=false]{hyperref}

\begin{document}

\title{Unsupervised Attention Based Instance Discriminative Learning for Person Re-Identification}

\author{Kshitij Nikhal\\
University of Nebraska-Lincoln\\
{\tt\small knikhal2@huskers.unl.edu}
\and
Benjamin S. Riggan\\
University of Nebraska-Lincoln\\
{\tt\small briggan2@unl.edu} 
}

\maketitle

\begin{abstract}
   Recent advances in person re-identification have demonstrated enhanced discriminability, especially with supervised learning or transfer learning. However, since the data requirements---including the degree of data curations---are becoming increasingly complex and laborious, there is a critical need for unsupervised methods that are robust to large intra-class variations, such as changes in perspective, illumination, articulated motion, resolution, etc. 
   Therefore, we propose an unsupervised framework for person re-identification which is trained in an end-to-end manner without any pre-training. Our proposed framework leverages a new attention mechanism that combines group convolutions to (1) enhance spatial attention at multiple scales and (2) reduce the number of trainable parameters by 59.6\%. Additionally, our framework jointly optimizes the network with agglomerative clustering and instance learning to tackle hard samples. We perform extensive analysis using the Market1501 and DukeMTMC-reID datasets to demonstrate that our method consistently outperforms the state-of-the-art methods (with and without pre-trained weights). 
  
\end{abstract}

\section{Introduction}
\begin{figure}[ht]
\centering
\includegraphics[width=0.85\linewidth]{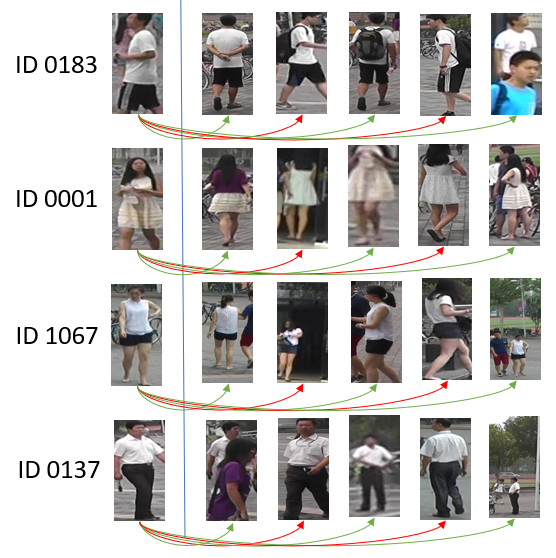}
\setlength{\belowcaptionskip}{-10pt}

   \caption{Example hard positive (\textcolor{Green}{green} arrows) and hard negative (\textcolor{Red}{red} arrows) examples for Re-ID.
   }
\label{fig:long}
\label{fig:onecol}
\end{figure}

Person re-identification (Re-ID) is the process of detecting identical persons to a query subject in frames from distributed cameras with non-overlapping field of views (FOVs). Given a query image~\cite{Zheng2015}, a video sequence~\cite{Wang2014} or text description~\cite{TextDescription}, Re-ID is used to track individuals in public spaces, such as airports, universities, malls, cities, etc. For example, Re-ID may be used to rapidly build a timeline for a person's whereabouts prior to a critical event (or crime), such as a mass shooting, bombing or other public threat, providing beneficial tools for local and federal law enforcement, military/government intelligence, surveillance and reconnaissance (ISR).

The primary challenge with Re-ID is detecting persons that match query subjects under significant variations in viewpoint, resolution, compression,
illumination, and occlusion.  Also, temporal variations such as changes in clothing, hair style, or accessories present additional challenges.

Despite initial success using hand crafted features~\cite{hcf1,hcf2} and metric learning~\cite{hcf3,hcf4}, current state-of-the-art~\cite{zhang2017alignedreid} Re-ID techniques are based on convolutional neural networks (CNNs), which, if supervised, require large-scale annotated (i.e., labeled) datasets to learn a robust embedding subspace. Comprehensive surveys on person Re-ID using hand crafted systems and recent deep learning techniques are presented in~\cite{SgongBook,Zheng2016PersonRP} and~\cite{deepreview2,ye2020deep}, respectively.

Annotating large-scale datasets for Re-ID, especially methods requiring multiple bounding boxes for each person, is very labor intensive, time consuming and cost prohibitive.
Therefore, in this paper, we propose a new unsupervised framework for Re-ID. 

Unsupervised methods commonly exploit pre-trained models or transfer learning, where parameters are assumed to generalize between two independent tasks, in order to improve the accuracy. For example, \cite{BUCLin} uses the Resnet50~\cite{he2015deep} architecture, pre-trained on ImageNet~\cite{Imagenet}, before fine-tuning the network in an unsupervised manner and~\cite{EUG} combines one-shot and unsupervised learning.
Although pre-training and transfer learning have been empirically shown to significantly enhance the performance of neural networks, it is not amenable for adapting parameters between significantly different domains or architectures.

In this paper, we propose a new framework that exploits ``soft'' attention maps (section~\ref{gam}), which are produced by efficiently and effectively combining group convolutions with channel-wise attention to enhance unsupervised spatial attention and to alleviate over-fitting by minimizing the number of trainable parameters. 

Additionally, instance discriminative learning and hierarchical clustering are jointly used to improve unsupervised learning with and without pre-training. 
Similar to supervised attention models, this joint optimization enables our attention mechanism to learn image regions that are most discriminative according to clustering metrics in an unsupervised manner (i.e., without labels).
We assign pseudo-labels to each training sample and learn to be robust to representations of the same instance under various perturbations (section~\ref{iwdl}). In successive stages, we tackle the hard positives, as shown in Figure~\ref{fig:long}, by gradually merging samples together using similarity scores (section~\ref{acl_loss_section}).

Our proposed framework achieves enhanced 
performance for unsupervised Re-ID (with and without pre-training) using the Market1501~\cite{market1501dataset} and DukeMTMC-reID~\cite{dukeimagedataset} datasets, beating the current state-of-the-art.


The contributions of this paper include:
\begin{compactitem}
\item a new grouped attention module (GAM),
\item the joint optimization of an instance discrimination loss (IDL) and agglomerative clustering loss (ACL),
\item extensive analysis for Re-ID using Market1501 and DukeMTMC-reID datasets,
\item ablation studies that analyze attention maps, embedding dimensionality, and number of filter groups.
\end{compactitem}

\section{Related Work}
{\bf Supervised Person Re-ID:} Supervised methods for Re-ID have been successful, in part, due to ubiquitous graphics processing units (GPUs), machine learning application programming interfaces (APIs), and large-scale datasets with annotations. However, supervised methods must ensure they are not over-fitting to any particular dataset.

Recently, AlignedReID~\cite{zhang2017alignedreid} achieved impressive performance on the Market1501~\cite{market1501dataset} dataset with a rank-1 accuracy of 94.4\% by jointly learning the global features with the local features. Zhou~\etal~\cite{zhou2019omniscale} proposed a new efficient architecture that achieves 94.8\% with 2.2 million parameters compared to the ResNet50~\cite{he2015deep} architecture with around 24 million parameters, reducing the possibility of overfitting. Zheng~\etal~\cite{zheng2019joint} use joint generative and discriminative learning and achieves an accuracy of 94.8\% on the Market1501 dataset. In this paper, we aim to minimize the gap between supervised and fully unsupervised performance.

\begin{figure*}[ht]
\centering
\includegraphics[width=0.95\linewidth]{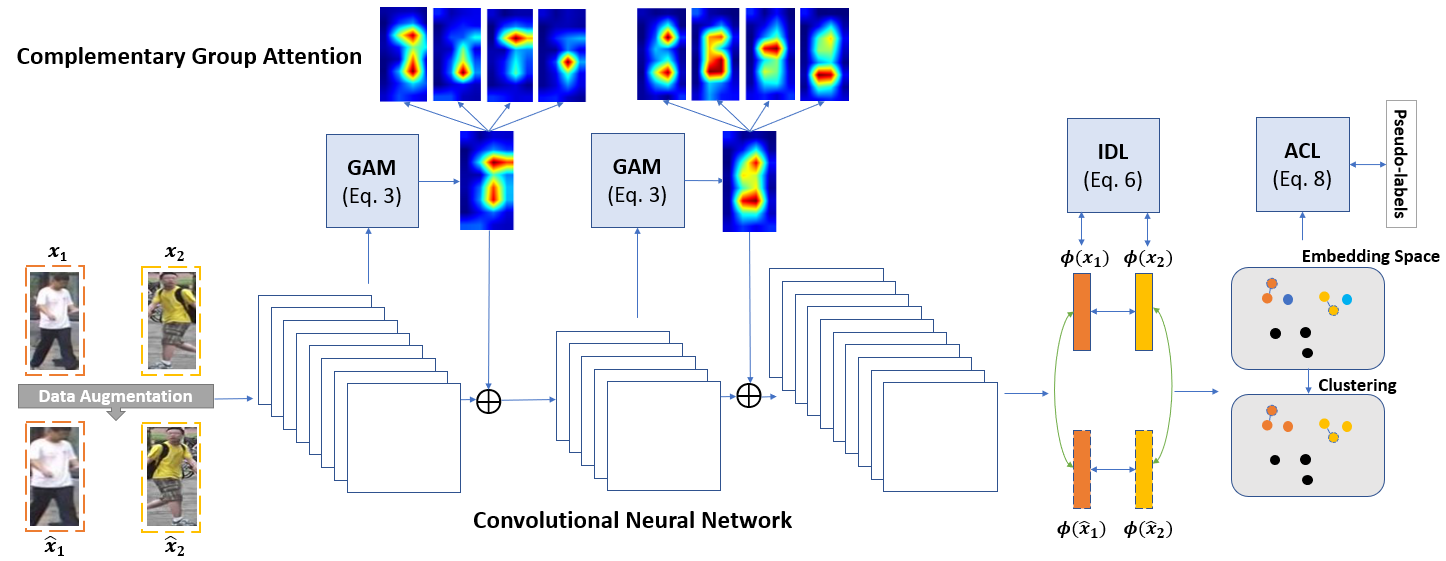}
    \caption{Our framework augments a network with Grouped Attention Modules (GAMs) at multiple scales and trains the network in a fully unsupervised manner using the instance discriminative loss (IDL) and agglomerative clustering loss (ACL). The complementary group attention maps helps generate better attention by learning filter relationships in a more structured way (e.g., the third filter group focuses on the top part only)}.
   \label{frameworkoverview}
\vspace{-4mm}
\end{figure*}

{\bf Unsupervised Person Re-ID:} Both dictionary learning and metric learning have been applied to unsupervised person Re-ID.
Kodirov~\etal~\cite{KodirovGraph} use a graph regularised dictionary learning algorithm with a robust L1-norm term to learn cross-view discriminative information. Yu~\etal~\cite{Yu_2020} learn a specific projection for each camera view based on asymmetric clustering. However, these features are not as discriminative as deeply learned features. Wu~\etal~\cite{EUG} focus on one-shot learning for video-based Re-ID to exploit unlabeled tracklets by progressively adding hard samples. Yu~\etal~\cite{yu2019unsupervised} learn a soft multi-label for each unlabeled person by comparing (and representing) the unlabeled person with a set of known reference persons from a labeled dataset. Xiao~\etal~\cite{xiao2016joint} propose an online instance matching loss function that maintains a lookup table of features from all the labeled identities, and compare distances between mini-batch samples and all the registered entries. Lin~\etal~\cite{BUCLin} use bottom up clustering (BUC) to optimize a CNN and balances cluster volume using a diversity term. 
However, several methods that claim to be unsupervised use some form of supervision like labeled supplementary datasets~\cite{yu2019unsupervised}, pre-trained ImageNet weights~\cite{BUCLin}, or one labeled tracklet for each identity~\cite{EUG}. In our work, we present experiments with and without pre-trained weights initialization. 



{\bf Attention Models:} Attention-based methods~\cite{li2018diversity,xu2018attentionaware} involve an attention mechanism to extract additional discriminative features. In comparison with pixel-level masks, attention regions can be regarded as an automatically learned high-level masks. This is analogous to a segmentation model that does not require any annotations and which is learned automatically. Li~\etal~\cite{li2018harmonious} proposed the Harmonious Attention CNN  (HA-CNN) model that combines the learning of soft pixel and hard regional attentions along with simultaneous optimization of feature representations. Zheng~\etal~\cite{zheng2018reidentification} proposed an attention-driven siamese network called the Consistent Attentive Siamese Network that uses identity labels as supervision. Wang~\etal~\cite{wang2017residual} proposed Residual Attention Network which uses an encoder decoder style attention module. Woo~\etal~\cite{woo2018cbam} proposed the Convolutional Block Attention Module (CBAM) that infers attention maps along channel and spatial dimensions sequentially.  Our attention mechanism is motivated by CBAM~\cite{woo2018cbam}, except that our proposed framework is more efficient, using substantially fewer parameters.

{\bf Unsupervised representation}: A classical approach for unsupervised representation learning is to perform clustering (\eg k-means).
Another popular method is to use auto-encoders~\cite{autoencoderbengio} to compress images into a latent representations, such that images may be optimally reconstructed.
Dosovitskiy~\etal~\cite{exemplarcnn} propose to discriminate among a set of surrogate classes, where the surrogate classes are formed by applying a variety of transformations to randomly sampled image patches. In~\cite{elis}, a softmax embedding variant is used where augmented samples should be classified as the same instance while other instances in the batch are considered as negative samples.

Unlike previous methods, our approach combines agglomerative clustering with an attention based CNN and pseudo-supervision in which neither labels nor pre-trained weights are used. 
Our approach differs from BUC by using a novel attention mechanism and an efficient architecture to produce discriminative representations associated with persons in the field of view. Moreover, our combination of IDL and ACL alleviates the need to use conventional pre-trained architectures.
Thus, our proposed approach is capable of being trained in a fully unsupervised manner. 

\begin{figure*}[ht]
\centering
\includegraphics[width=0.8\linewidth]{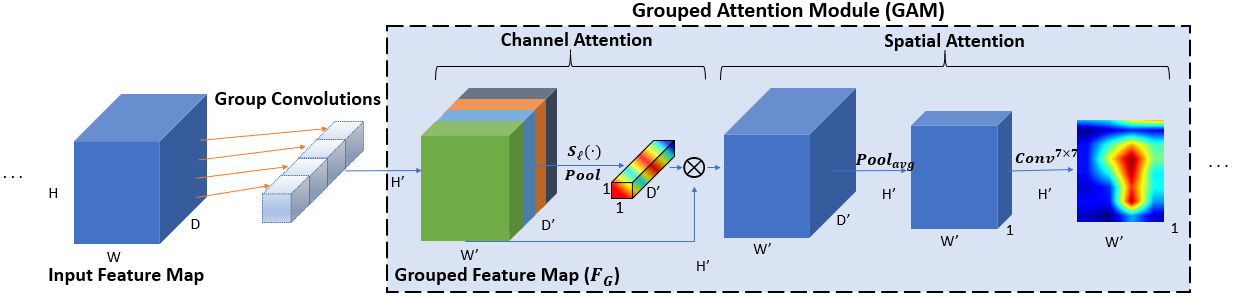}

  \caption{Overview of our Group Attention Module (GAM), combining group-based channel attention and spatial attention.  }
\label{gamarch}
\vspace{-4mm}
\end{figure*}

\section{Preliminaries}

Given an unlabeled training set $X = \{x_1, x_2, ...x_n\}$, the goal is to learn $\phi(x_{i};\theta)$---a mapping parameterized by $\theta$ used to extract features from an image $x_i$.
This mapping is then applied to the gallery set $X^g = \{x^g_1, x^g_2, ...x^g_{ng}\}$ and query set $X^q = \{x^q_1, x^q_2, ...x^q_{nq}\}$. The gallery set can be considered the test set or the total collection of detections in the database. Representations of the query images, $\phi(x^q_i;\theta)$ are used to search the gallery set to retrieve the most similar matches 
to $x_i^q$ according to Euclidean distance between the query and gallery embeddings, $d(x_q, x_g) = \| \phi(x_q; \theta) - \phi(x_g; \theta)\|$, where a smaller distance implies increased similarity between the images. 
Ideally, the top-$k$ (for $k$ equal to 1, 5, or 10) matches returned will correspond to the same identity as that from the query image.


\section{Methodology}
Our hybrid framework for fully unsupervised Re-ID (Figure~\ref{frameworkoverview})
consists of three main components: Grouped Attention Module (GAM), Instance Discriminative Loss (IDL), and Agglomerative Clustering Loss (ACL). 

Our GAM, which uses reduced filter group co-dependence \cite{srivastava2014dropout} in conjunction with compact channel and spatial attention, yields more accurate (and efficient) attention maps. Similar to \cite{ioannou2016deep, krizhevsky2012imagenet}, our GAM consists of convolutional layers with filter groups that consistently learn complementary attention maps. Thus, more discriminative representations are generated for enhancing clustering capabilities by focusing on persons' appearances rather than background/clutter, which is observed by the increased accuracy between GAM and CBAM in section~\ref{sec:quantitative}.\\ 
\indent IDL and ACL have complementary objectives in regards to unsupervised Re-ID. 
IDL plays an important role by making the network invariant to different cross-camera views by maximizing the similarity measure between the original images and augmentations (e.g., zoomed/flipped). Unlike IDL, ACL aims to iteratively merged clusters and optimize similarity between image representations (enhanced by GAM and IDL) and associated clusters.  Therefore, we jointly optimize both losses to more effectively learn  discriminative representations for unsupervised person Re-ID. 
 
Each component is discussed in subsections~\ref{gam}--\ref{acl_loss_section}.

\subsection{Grouped Attention Module (GAM)} \label{gam}

The proposed attention mechanism works by summarizing activations of previous layers to emphasize ``important'' regions for subsequent layers. The attention map is inferred from an intermediate feature map to generate attention aware features. The attention module is split into two attention maps: channel and spatial attention (Figure~\ref{gamarch}).

{\bf Channel Attention}: This module exploits the inter-channel relationship of the features, focusing on the important features in the input image. The intermediate feature map is aggregated to remove spatial information and retain the channel information only. It is aggregated by using a sub-network to infer channel-wise attention. In our approach, we use average pooling layers to remove the spatial information followed by a linear layer. 
Let $F$ be an intermediate feature map of dimension $\mathbb{R}^{C \times H \times W}$ that is input to the attention module. The channel attention is 

\begin{equation}
A_c(F) = \sigma(S_\ell(Pool_{avg}(F))),
\label{eq:one}
\end{equation}
where $S_\ell$ is a fully connected layer and $\sigma$ is the sigmoid activation function.


The intermediate convolutional layers, producing the input $F$ in Eq.~\ref{eq:one}, are replaced with group convolutions, whose feature maps are denoted as $F_G$.
This significantly reduces the parameters while improving the attention maps (section~\ref{sec:ablation}). 
We use 4 filter groups which reduces the parameters from 26 million parameters to 10.5 million parameters, hence making the network efficient by reducing 59.6\% of the parameters compared to the Resnet50~\cite{he2015deep}. So the channel attention can now be stated as,

\begin{equation}
A_c(F_G) = \sigma(S_\ell(Pool_{avg}(F_G))).
\end{equation}

{\bf Spatial Attention Module}: The spatial attention module is used to learn the pixels that contribute the most toward the network's inference.
For every spatial index $(u,v)$ for $F_G$, $A_c$ and $F_G(u,v)$ are multiplied in an element-wise manner.  For compactness, we denote the procedure as $A_c \otimes F_G$. This new channel attention enhanced feature map is passed to the spatial attention module.
Average pooling is applied along the channel axis, which is shown to be effective in highlighting informative regions~\cite{zagoruyko2016paying}.
Then, we apply a $7 \times 7$ convolutional layer to generate a spatial attention map that encodes 
whether pixels are emphasized or deemphasized.

\begin{equation}
A_s(A_c(F_G)) = \sigma((Conv^{7 \times 7}(Pool_{avg}(A_c(F_G)\times F_G)))),
\end{equation}
where $A_s(\cdot) \in \mathbb{R}^{1 \times H \times W}$.

We refer to this sequential attention inference with group convolutions as the grouped attention module (GAM). Despite not being explicitly trained in a supervised manner, our network learns to focus on discriminative image regions. Specifically, as shown in section~\ref{exp}, the attention maps and feature representations are improved.

\subsection{Instance Discriminative Loss (IDL)} \label{iwdl}

Ideally, the objective is to retrieve genuine correspondences to the query subject, which implies that we desire genuine and imposter similarity/dissimilarity score distributions that are maximally separable.
Therefore, our proposed framework, including GAMs, is trained such that two feature embeddings from imagery acquired across multiple camera views for the same subject are sufficiently close in terms of Euclidean distance and embeddings from different subjects produce higher dissimilarity scores.

Cross-camera views can be imitated to a certain extent with data augmentations. We use various augmentation techniques like random crops, zoom, horizontal flips and occlusions to approximate cross-view variations, and we use pseudo-supervision to classify instances (with unknown identities) to be the same as their augmentations.

Thus, we minimize the difference between an instance and its augmented version while maximizing the difference between other instances.
This can be modeled as a binary classification problem. Particularly, for sample $x_i$, the augmented sample $\hat{x_{i}}$ is classified as instance $i$ while other samples, $x_j$ for $j\ne i$ are not classified as instance $i$.
The probability of an augmentation, $\hat{x_{i}}$ being classified the same as image $x_i$ is

\begin{equation}
     P(i|\hat{x_{i}}) = \frac{\exp(\phi(x_i;\theta)^T\phi(\hat{x_i};\theta) /\tau)}{\sum_k \exp(\phi(x_k;\theta)^T\phi(\hat{x_i};\theta) /\tau)}, 
     \label{pos}
 \end{equation}
 where $\tau$ is a temperature parameter that controls the softness of the probability distribution~\cite{hinton2015distilling}.
 Since the embeddings are L2 normalized, maximizing the numerator in Eq.~\ref{pos} implies increasing the cosine similarity between $\phi(x_i;\theta)$ and $\phi(\hat{x_i};\theta)$. Thus, maximizing Eq.~\ref{pos} encourages instances to be robust to cross-camera views using data augmentations. 
The probability of $x_j$ being classified as instance $i$ is 
 
 \begin{equation}
     P(i|x_{j}) = \frac{\exp(\phi(x_i;\theta)^T\phi(x_j;\theta) /\tau)}{\sum_k \exp(\phi(x_k;\theta)^T\phi(x_j;\theta) /\tau)}. 
     \label{neg}
 \end{equation}
 
 Therefore, we want to maximize Eq.~\ref{pos} and minimize Eq.~\ref{neg} which is equivalent to minimizing the negative log likelihood
 
 \begin{equation}
 J_{idl} = -\sum_{i} \log P(i|\hat{x_{i}}) - \sum_{i}\sum_{j \ne i} \log(1 - P(i|x_{j}).
 \label{costinstance}
\end{equation}

\renewcommand{\subtablename}{Table}
\renewcommand*{\thesubtable}{\arabic{table}}
\makeatletter
\renewcommand{\p@subtable}{}
\makeatother

\renewcommand{\thefootnote}{\fnsymbol{footnote}}

\begin{table*}[htb]
\centering
\captionsetup[subtable]{labelformat=simple, labelsep=colon, font=large}

\setlength\tabcolsep{2pt} 
\begin{subtable}{0.52\textwidth}
\caption{Market1501 results using pretrained weights.}
\label{marketdataset1}
\begin{tabular*}{\textwidth}{|c|c|c|c|c|c| } 
\cline{1-6}
Method & Label & rank-1 & rank-5 & rank-10 & mAP \\ 
\cline{1-6}
 BOW~\cite{market1501dataset} & ImageNet & 35.8\% & 52.4\% & 60.3\% & 14.8\% \\
 OIM~\cite{xiao2016joint} & ImageNet & 38.0\% & 58.0\% & 66.3\% & 14.0\% \\ 
 UMDL~\cite{lv2018unsupervised} & Transfer & 34.5\% & 52.6\% & 59.6\% & 12.4\% \\ 
 PUL~\cite{fan2017unsupervised} & Transfer & 44.7\% & 59.1\% & 65.6\% & 20.1\% \\ 
 EUG~\cite{EUG} & OneShot & 49.8\% & 66.4\% & 72.7\% & 22.5\% \\ 
 SPGAN~\cite{deng2017imageimage} & Transfer & 58.1\% & 76.0\% & 82.7\% & 26.7\% \\ 
 TJ-AIDL~\cite{wang2018transferable} & Transfer & 58.2\% & - & - & 26.5\% \\ 
 BUC~\cite{BUCLin} & ImageNet & 61.9\% & 73.5\% & 78.2\% & 29.6\% \\ 
 \cline{1-6}
 IDL\footnotemark[1] & ImageNet & 39.7\% & 56.8\% & 64.7\% & 15.3\% \\ 
 IDL + ACL\footnotemark[1] & ImageNet & 56.9\% & 74.8\% & 80.6\% & 30.1\% \\ 
 IDL + ACL + CBAM\footnotemark[1] & ImageNet & 63.3\% & \textbf{81.3}\% & \textbf{86.3}\% & 35.0\% \\
 ACL + GAM \footnotemark[1] & ImageNet & 57.3\% & 75.3\% & 81.9\% & 28.3\% \\ 
 IDL + ACL + GAM\footnotemark[1] & ImageNet & \textbf{63.9}\% & 78.8\% & 85.1\% & \textbf{35.7}\% \\ 
 
\cline{1-6}
\end{tabular*}
\footnotetext{*~Our approach / implementation.}
\end{subtable}
~
\setlength\tabcolsep{2pt} 
\addtocounter{table}{1} 
\begin{subtable}{0.44\textwidth}
\caption{Market1501 results (no pretrained weights).}
\label{marketdataset2}
\begin{tabular*}{\textwidth}{|c|c|c|c|c| } 
\cline{1-5}
 Method & rank-1 & rank-5 & rank-10 & mAP \\ 
\cline{1-5}
 BUC\footnotemark[1]~\cite{BUCLin} & 10.7\% & 21.7\% & 27.8\% & 3.1\% \\
 gBiCov~\cite{xiong2014person} & 8.28\% & - & - & 2.23\% \\
 HistLBP~\cite{ma2014covariance} & 9.62\% & - & - & 2.72\% \\
 LOMO~\cite{liao2015person} & 26.07\% & - & - & 7.75\% \\
 BOW+MultiQ~\cite{market1501dataset} & 42.64\% & - & - & 18.68\% \\
 \cline{1-5}
 IDL\footnotemark[1] & 25.5\% & 42.4\% & 51.0\% & 9.1\% \\ 
 IDL + ACL\footnotemark[1] & 48.2\% & 68.0\% & 76.2\% & 24.3\% \\ 
 IDL + ACL + CBAM\footnotemark[1] & 48.5\% & 67.2\% & 74.3\% & 25.6\% \\ 
 IDL + ACL + A-CBAM \footnotemark[1] & 45.2\% & 63.1\% & 70.6\% & 21.7\% \\
 ACL + GAM \footnotemark[1] & 44.6\% & 63.5\% & 71.6\% & 20.3\% \\ 
 IDL + ACL + GAM\footnotemark[1] & \textbf{53.6}\% & \textbf{71.9}\% & \textbf{79.4}\% & \textbf{29.5}\% \\ 
\cline{1-5}
\end{tabular*}
\footnotetext{*~Our approach / implementation.A-CBAM - Adjusted CBAM i.e. parameters are controlled to match GAM.}
\end{subtable}
\vspace{-5mm}
\end{table*}

\subsection{Agglomerative Clustering Loss (ACL)} \label{acl_loss_section}

To improve the discriminative capabilities, our framework incorporates agglomerative clustering, which is a form of hierarchical clustering that successively merges instances into clusters in a bottom up manner.
Each cluster center $M_{\beta}$ for $\beta \in \{1 \dots |M|\}$ is used to form a memory bank, $M$, where $|M|$ denotes the size of the memory bank (i.e., the number of clusters).  Initially, $|M|=n$, meaning that all training instances $x_i$ for $i = 1 \dots n$ are their own singleton clusters. However, as instances merge (i.e, $|M|<n$), non-singleton clusters are formed.


Let $\beta_i$ denote the cluster label corresponding to $x_i$ for $i=1\dots n$.
The probability that image $x_i$ belongs to a cluster $\beta_i$ is given by
 \begin{equation}
     P(\beta_i|x_{i}) = \frac{\exp(M_{\beta_i}^T \phi(x_i;\theta)  /\tau)}{\sum_k \exp(M_{\beta_k}^T\phi(x_i;\theta) /\tau)}.
     \label{aggl}
 \end{equation}
At each learning iteration, the representation $\phi(x_i;\theta)$ and the parameters $\theta$ are optimized by stochastic gradient descent. Then, the optimized representation is used to update the memory bank $M$.
Therefore, the objective function is to minimize the negative log likelihood as,
\begin{equation}
J_{acl} = - \sum^n_{i=1} \log P(\beta_i|x_i).
\label{eq:acl}
\end{equation}

Successively, each singleton cluster is then merged together according to a dissimilarity metric using the feature representation. The number of clusters $N_C$ to merge is a hyper-parameter and can be dynamically changed during training. We use the Euclidean distance metric:

\begin{equation}
    d_0(\phi(x_i),\phi(x_j)) = \sqrt{\phi_{ij}^T\phi_{ij}},
\end{equation}
where we dropped the parameter, $\theta$, for compactness and $\phi_{ij}=\phi(x_i) - \phi(x_j)$.

To avoid mode collapse, we use a balancing term in the similarity measure as described in~\cite{BUCLin}

\begin{equation}
d(\phi(x_i),\phi(x_j)) = d_0 + \lambda(|M_i| + |M_j|)
\end{equation}
where $\lambda$ is the weighting term to balance the impact of the balancing term. This helps in merging small clusters while also merging large clusters that are very similar. 

Our proposed framework uses both the IDL (Eq.~\ref{costinstance}) and ACL (Eq.~\ref{eq:acl}) to train a network using GAMs. The total loss is expressed as 

\begin{equation}
    J_{total} = J_{idl} + J_{acl}.
\end{equation}


\section{Experiments} \label{exp}
In this section, we describe the extensive experimental analysis and summarize the results using two established Re-ID benchmark datasets. First, we describe the datasets and implementation details for purposes of reproducibility. Then, we provide quantitative and qualitative analysis, comparing our proposed unsupervised framework with recent unsupervised methods (with and without pre-training).
Lastly, we present ablation studies that analyze attention maps, embedding dimensions, and number of filter groups. 

\subsection{Datasets}
For experimental analysis, we employed two widely used image-based Re-ID datasets: the Market1501 and DukeMTMC-reID.

The \textbf{Market1501}~\cite{market1501dataset} dataset uses imagery from six cameras in front of the Tsinghua University campus. This dataset contains 32,668 bounding boxes of 1,501 identities. The training set consists of 12,936 images of 751 identities and the testing set contains 19,732 of 750 identities. 
The dataset uses the Deformable Part Model~\cite{dpm} as opposed to hand drawn bounding boxes. This introduces misalignment, part missing and false positives which reflects a realistic setting. The dataset is collected in an open system, where each identity has multiple images under each camera.

The \textbf{DukeMTMC-reID}~\cite{dukeimagedataset} dataset is a subset of the DukeMTMC~\cite{dukevideodataset} dataset for image-based Re-ID which is similar to the format of the Market1501 dataset. There are 1,404 identities that appear in more than two cameras and 408 distractors who appear only in one camera. The dataset is split by randomly selecting 702 identities as the training set and 702 identities as the testing set. In the testing set, one image is picked for each identity as the query image while the others are put in the gallery set. As a result, a total of 16,552 images are available for training and 19,889 images are available for testing.

\textbf{Evaluation:} We followed the evaluation protocols described for the respective datasets, where the query and gallery samples were captured by different cameras. Then, given a query image sequence, all gallery items were assigned a similarity score and were ranked according to their similarity with the query. The
search process was performed in a cross-camera mode, i.e., relevant images captured in the same camera as the query were not considered.

For quantitative analysis, we used the rank-$k$ accuracy (for $k=1,5,$ and $10$) and mean average precision (mAP) to compare our proposed framework (and components) with other methods.  The rank-$k$ accuracy is derived from the cumulative matching characteristic (CMC) curve, where a match is considered to be correct if a true match (according to the ground truth set) is returned within the top-$k$ most similar matches.  The average precision (AP) is computed as the area under the precision-recall curve, and the mAP is the average over all queries.  A key difference between rank-$k$ accuracy and mAP is that rank-$k$ considers the most similar match scores and mAP considers the separability between genuine and imposter score distributions.


\subsection{Implementation Details}
The CNN used in our experiments to evaluate the proposed framework 
was based on the Resnet50 architecture. Resnet50 is composed of 16 bottleneck blocks: $Bottleneck(x)=Relu(\mathcal{F}(x) + x)$, where
$\mathcal{F}(x)=Conv \circ BN \circ Conv \circ BN\circ Relu \circ Conv \circ BN \circ Relu(x)$ and $Conv$, $BN$, and $Relu$ represent convolution, batch normalization, and rectified linear unit layers, respectively.
However, with our GAM, the bottleneck layers are $Relu(GAM \circ \mathcal{F}(x) + x)$ and all $Conv$ layers in $\mathcal{F}(x)$ are replaced with group convolutions using 4 filter groups.

All proposed methods are optimized using stochastic gradient descent (SGD) with weight decay of 0.9 and with a batch size of 32. Initially, the effective learning rate is set to 0.01 when layers are initialized with pre-trained weights and 0.1 otherwise. After 25 epochs, the learning rate is reduced by a factor of 10. The merging percent of clusters is set to 4\%. 
All input images were resized to $256 \times 128$ (H $\times$ W) and augmented using random horizontal flips, random contrast change, random zoom and random crop. The temperature parameter $\tau$ in Eqs.~\ref{pos}, \ref{neg} and \ref{aggl} is set to 0.1.

In the subset of experiments in which we initialized using pre-trained weights, we pre-trained our modified Resnet50 architecture on ImageNet to provide a fair comparison with other architectures that were either pre-trained on ImageNet in a similar fashion or used some alternative discriminative pre-training.  However, for completeness, we also compare our proposed framework with other methods without any discriminative pre-training.

\begin{table*}[htb]
\centering
\captionsetup[subtable]{labelformat=simple, labelsep=colon, font=large}

\setlength\tabcolsep{2pt} 
\begin{subtable}{0.52\textwidth}
\caption{DukeMTMC-reID results with pretrained weights.}
\label{dukedataset1}
\begin{tabular*}{\textwidth}{|c|c|c|c|c|c| } 
\cline{1-6}
Method & Label & rank-1 & rank-5 & rank-10 & mAP \\ 
\cline{1-6}
 BOW~\cite{market1501dataset} & ImageNet & 17.1\% & 28.8\% & 34.9\% & 8.3\% \\
 OIM~\cite{xiao2016joint} & ImageNet & 24.5\% & 38.8\% & 46.0\% & 11.3\% \\ 
 UMDL~\cite{lv2018unsupervised} & Transfer & 18.5\% & 31.4\% & 37.6\% & 7.3\% \\ 
 PUL~\cite{fan2017unsupervised} & Transfer & 30.4\% & 46.4\% & 50.7\% & 16.4\% \\ 
 EUG~\cite{EUG} & OneShot & 45.2\% & 59.2\% & 63.4\% & 24.5\% \\ 
 SPGAN~\cite{deng2017imageimage} & Transfer & 46.9\% & 62.6\% & 68.5\% & 26.4\% \\ 
 TJ-AIDL~\cite{wang2018transferable} & Transfer & 44.3\% & - & - & 23.0\% \\ 

 BUC~\cite{BUCLin} & ImageNet & 40.4\% & 52.5\% & 58.2\% & 22.1\% \\ 
 \cline{1-6}
 IDL\footnotemark[1] & ImageNet & 25.3\% & 40.4\% & 46.9\% & 11.2\% \\ 
 IDL + ACL\footnotemark[1] & ImageNet & 43.9\% & 59.6\% & 66.1\% & 23.7\% \\ 
 IDL + ACL + CBAM\footnotemark[1] & ImageNet & 46.0\% & 62.4\% & 69.1\% & 26.0\% \\
 ACL + GAM \footnotemark[1] & ImageNet & 42.7\% & 59.6\% & 66.0\% & 23.6\% \\ 
 IDL + ACL + GAM\footnotemark[1] & ImageNet & \textbf{47.2\%} & \textbf{63.8\%} & \textbf{69.8\%} & \textbf{28.1\%} \\ 
 
\cline{1-6}
\end{tabular*}
\footnotetext{*~Our approach / implementation.}
\end{subtable}
\quad
\setlength\tabcolsep{2pt} 
\addtocounter{table}{1} 
\begin{subtable}{0.44\textwidth}
\caption{DukeMTMC-reID results (no pretrained weights).}
\label{dukedataset2}
\begin{tabular*}{\textwidth}{|c|c|c|c|c| } 
\cline{1-5}
 Method & rank-1 & rank-5 & rank-10 & mAP \\ 
\cline{1-5}
 BUC\footnotemark[1]~\cite{BUCLin} & 5.1\% & 9.6\% & 11.7\% & 1.6\% \\
 BOW~\cite{market1501dataset} & 17.1\% & 28.8\% & 34.9\% & 8.3\% \\
 UMDL~\cite{lv2018unsupervised} & 18.5\% & 31.4\% & 37.6\% & 7.3\% \\ 
 \cline{1-5}
 IDL\footnotemark[1] & 10.6\% & 22.2\% & 28.7\% & 4.3\% \\ 
 IDL + ACL\footnotemark[1] & 30.6\% & 47.4\% & 53.7\% & 17.0\% \\ 
 IDL + ACL + CBAM\footnotemark[1] & 34.1\% &  \textbf{53.3}\% & 60.4\% & 21.2\% \\ 
 IDL + ACL + A-CBAM \footnotemark[1] & 30.4\% & 48.1\% & 55.9\% & 17.7\% \\
 ACL + GAM \footnotemark[1] & 28.7\% & 45.6\% & 53.2\% & 15.6\% \\ 
 IDL + ACL + GAM\footnotemark[1] & \textbf{36.2}\% & 53.2\% & \textbf{60.9}\% & \textbf{22.9}\% \\ 
\cline{1-5}
\end{tabular*}
\footnotetext{*~Our approach / implementation.A-CBAM - Adjusted CBAM i.e. parameters are controlled to match GAM.}
\end{subtable}
\vspace{-5mm}

\end{table*}

\subsection{Quantitative Analysis}
\label{sec:quantitative}
In the following experiments, we compare our proposed framework with bottom up clustering (BUC) \cite{BUCLin} and several recent methods. We provide the results of fair comparisons for two scenario: with and without pre-trained weights.

Using the Market1501 dataset, Table~\ref{marketdataset1} 
shows the rank-1, rank-5, and rank-10 Re-ID accuracy and mAP under the
unsupervised scenario with pre-trained weights. Our proposed framework, using IDL and ACL to train the Resent50 architecture with our GAM, achieved the best rank-1 accuracy and mAP. Whereas, our approach using CBAM achieved the best rank-5 and rank-10 accuracy. Note that, in this case, we improved upon the state-of-the-art rank-1 accuracy and mAP reported in \cite{BUCLin} by 2.0\% and 6.1\%, respectively. 

Table~\ref{marketdataset2} shows the rank-1, rank-5, and rank-10 Re-ID accuracy and mAP for the Market1501 dataset under the challenging fully unsupervised scenario (i.e., no pre-trained weights are used).  The table shows that our proposed framework achieved a 36.3\% and 26.4\% improvement over \cite{BUCLin} in rank-1 accuracy and mAP.  Notice that the difference in performance between Tables~\ref{marketdataset1} and \ref{marketdataset2} for BUC \cite{BUCLin} is 51.2\% for rank-1 accuracy and 26.5\% for mAP.  Whereas, the difference for our proposed framework is significantly reduced, achieving a separation of 10.3\% in rank-1 accuracy and 6.2\% in mAP. Thus, our approach minimizes the gap between unsupervised learning with and without discriminative pre-training.  

Similarly, for the DukeMTMC-reID dataset, we obtained the best performance among all the compared methods with rank-1 of 47.2\% and mAP of 28.1\% (Table~\ref{dukedataset1}). For the fully unsupervised scenario shown in Table~\ref{dukedataset2}, we achieved a rank-1 performance of 36.2\% and mAP of 22.9\% which surpasses the state-of-the-art by 31.1\% and 21.3\%, respectively. Moreover, our methods also outperform all methods that leverage supplementary datasets or one shot labeled examples.



 

\begin{figure*}[htb]
\centering
\includegraphics[width=0.8\linewidth]{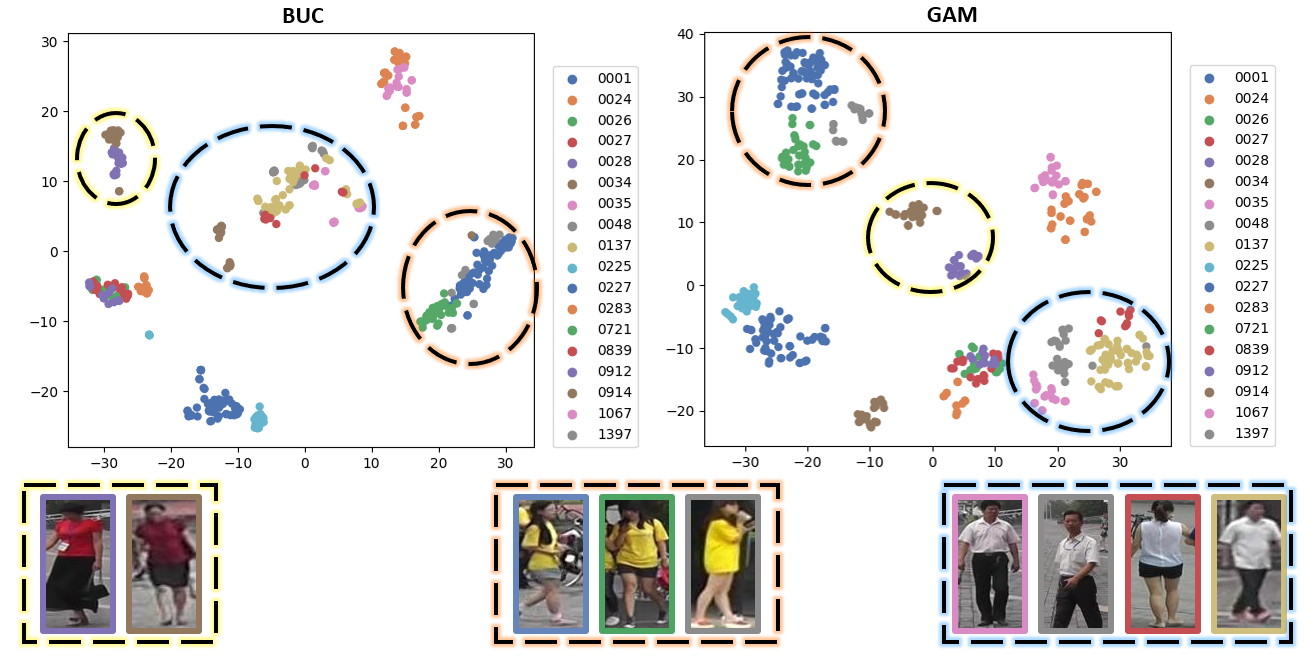}

   \caption{
   Examples of similar subjects that are closely clustered using t-SNE visualization are circled to show how our framework provides increased separability between different subjects over BUC~\cite{BUCLin}.}

\label{tsnediag}
\vspace{-4mm}

\end{figure*}

\subsection{Qualitative Analysis}
To evaluate our algorithm, t-distributed stochastic neighbour embedding (t-SNE)~\cite{maaten2008visualizing} is used to visualizes representations from BUC~\cite{BUCLin} and our proposed framework using the same data and perplexity to compare cluster quality. We pick 18 specific subjects that have hard positives and hard negatives. Figure~\ref{tsnediag} shows that our approach has better separability and structure of clusters compared with \cite{BUCLin}, 
which agrees with the quantitative results from (section~\ref{sec:quantitative}).

\subsection{Ablation Studies}
\label{sec:ablation}
Next, we summarize the results of a few ablation studies that consider the effects of GAM, embedding dimensionality, and number filter groups.

\begin{figure}[h]

\centering
\includegraphics[width=0.85\linewidth]{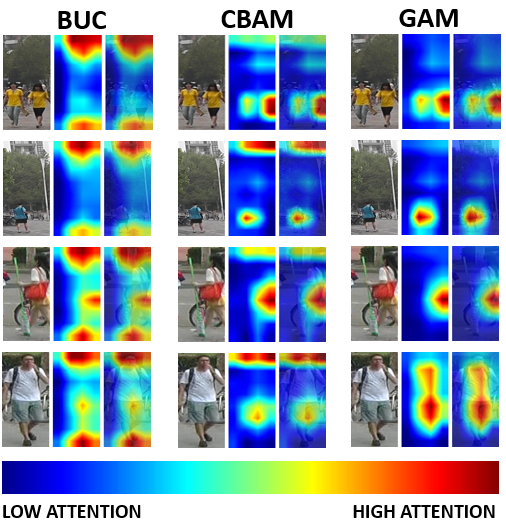}
   
   \caption{Comparison between BUC, CBAM, and GAM attention maps, with original image (left), attention map (middle), superimposed attention map and image (right). 
   }

\label{attnmap_dig}
\vspace{-5mm}
\end{figure}

\textbf{Attentions Maps:} We compare the activation maps at the final residual block for BUC~\cite{BUCLin}, CBAM~\cite{woo2018cbam}, and our proposed GAM. Figure~\ref{attnmap_dig} shows that the lacks of attention modeling prevents BUC from capturing the important details which should contribute to the representation. CBAM mitigates this problem but still has distracting activations associated with clutter rather than Re-IDs (\eg top of images). Our proposed GAM is able to minimize activations due to clutter and provide enhanced attention for Re-ID.

\textbf{Embedding Size:} We compare the feature embedding size on the DukeMTMC-reID dataset. The rank-1 accuracy seems to increase as embedding size increases (to a point). However, we need compact representations especially in applications like Re-ID.  Although Table~\ref{tab:embedding} shows that a 4096 dimensional embeddings achieves the best performances, either 1024 or 2048 dimensional embeddings may better balance performance and efficiency.  Making the embedding size even larger than 4096 dimensions had no significant effect on the accuracy. 

\begin{table}[h]
\centering
\caption{Results on the DukeMTMC-reID dataset with varying embedding sizes without pre-trained weights}
\label{tab:embedding}
\begin{tabular}{ |c|c|c|c|c| } 
\hline
 Feature Size & rank-1 & rank-5 & rank-10 & mAP \\ 
 \hline
 512 & 35.2\% & 52.5\% & 58.8\% & 22.4\% \\
 1024 & 35.6\% & 53.3\% & 61.9\% & 22.8\% \\  
 2048 & 36.2\% & 53.2\% & 60.9\% & 22.9\% \\
 4096 & \textbf{38.9}\% & \textbf{56.5}\% & \textbf{63.1}\% & \textbf{24.4}\% \\ 
 
 \hline
\end{tabular}
\vspace{-4mm}
\end{table}

\textbf{Number of Filter Groups:} Table~\ref{tab:filtergroup} compares fully unsupervised Re-ID performance when varying the number of filter groups for GAM on the Market1501 dataset. Increasing the number of filter groups led to better performance while significantly reducing parameters. Notice that group convolutions with 8 filter groups seemed to achieve the best rank-1 accuracy and mAP. However, increasing beyond 8 filter groups degraded performance, which is likely due to eliminating too many parameters.

\begin{table}[h]
\centering
\caption{Results on the Market1501 dataset with varying number of filter groups without pre-trained weights}
\label{tab:filtergroup}
\begin{tabular}{ |c|c|c|c|c| } 
\hline
 Filter Groups & rank-1 & rank-5 & rank-10 & mAP \\ 
 \hline
 2 & 51.1\% & 69.2\% & 76.0\% & 26.3\% \\
 4 & 53.6\% & 71.9\% & \textbf{79.4\%} & 29.5\% \\  
 8 & \textbf{55.0\%} & \textbf{72.5\%} & 79.2\% & \textbf{30.0\%} \\
 16 & 54.4\% & 71.2\% & 78.9\% & 28.6\% \\ 
 
 \hline
\end{tabular}
\end{table}

\section{Conclusion}
Unsupervised person Re-ID is especially difficult  under  significant  variations  in viewpoint,  resolution,  compression,  illumination,  and  occlusion. In this paper, we studied the effects of group based attention, instance discrimination, and agglomerative clustering, where we demonstrated that our framework composed of all three provided state-of-the-art Re-ID performance on the Market1501 and DukeMTMC-reID datasets.  Specifically, we showed that whether or not the network is pre-trained on ImageNet, our proposed framework provided the best Re-ID performance.  Most importantly, we showed that by providing a more discriminative unsupervised method without initializing with pre-trained weights, we significantly reduced the gap between supervised and unsupervised methods from greater than 80\% down to around 40\% rank-1 accuracy for the Market1501 and DukeMTMC-reID datasets.  The impacts of closing this gap are that (1) novel architectures (not pre-trained on ImageNet or similar datasets) may be more efficiently developed, without having to pre-train every network modification, and (2) new large-scale data collections may be simplified by reducing the amount of laborious, time consuming, and cost prohibitive data annotation process. 

{\small
\bibliographystyle{ieee_fullname}
\bibliography{gam}
}

\end{document}